\PassOptionsToPackage{dvipsnames,table}{xcolor}

\documentclass{vgtc}                          

\usepackage[svgnames]{xcolor}  

\graphicspath{{figures/}{pictures/}{images/}{./}}

\usepackage{times}

\usepackage{mathptmx}
\usepackage{amsmath}
\usepackage{amssymb}
\usepackage{enumitem}
\usepackage{tabularx}
\usepackage{booktabs}

\onlineid{1344}
\vgtccategory{Research}

\newcommand{\new}[1]{#1}                       

\AtEndPreamble{%
  \hypersetup{
    pdfsubject={IEEE ISMAR 2026 Conference Paper},
    pdfauthor={Chenfeng Gao and Taeyoung Yeon and Sungheon Park and Vasco Xu and Anish Prabhu and Mar Gonzalez-Franco and Karan Ahuja}}
}

\title{FlowAvatar: Real-Time Full-Body Avatars from Sparse Egocentric Inputs on Consumer XR Devices}

\author{%
  Chenfeng Gao\textsuperscript{1}\thanks{e-mail: chenfenggao2029@u.northwestern.edu, \{taeyoungyeon, sungheon.park, kahuja\}@northwestern.edu. C.~Gao and T.~Yeon contributed equally to this work.}, %
  Taeyoung Yeon\textsuperscript{1}, %
  Sungheon Park\textsuperscript{1}, %
  Vasco Xu\textsuperscript{2}\thanks{e-mail: vascoxu@uchicago.edu}, %
  Anish Prabhu\textsuperscript{3}\thanks{e-mail: \{anishprabhu, margon\}@google.com}\\[2pt]%
  Mar Gonzalez-Franco\textsuperscript{3}, %
  Karan Ahuja\textsuperscript{1}%
}
\affiliation{\scriptsize
  \textsuperscript{1}Northwestern University \quad
  \textsuperscript{2}University of Chicago \quad
  \textsuperscript{3}Google}

\teaser{
  \centering
  \includegraphics[width=\linewidth]{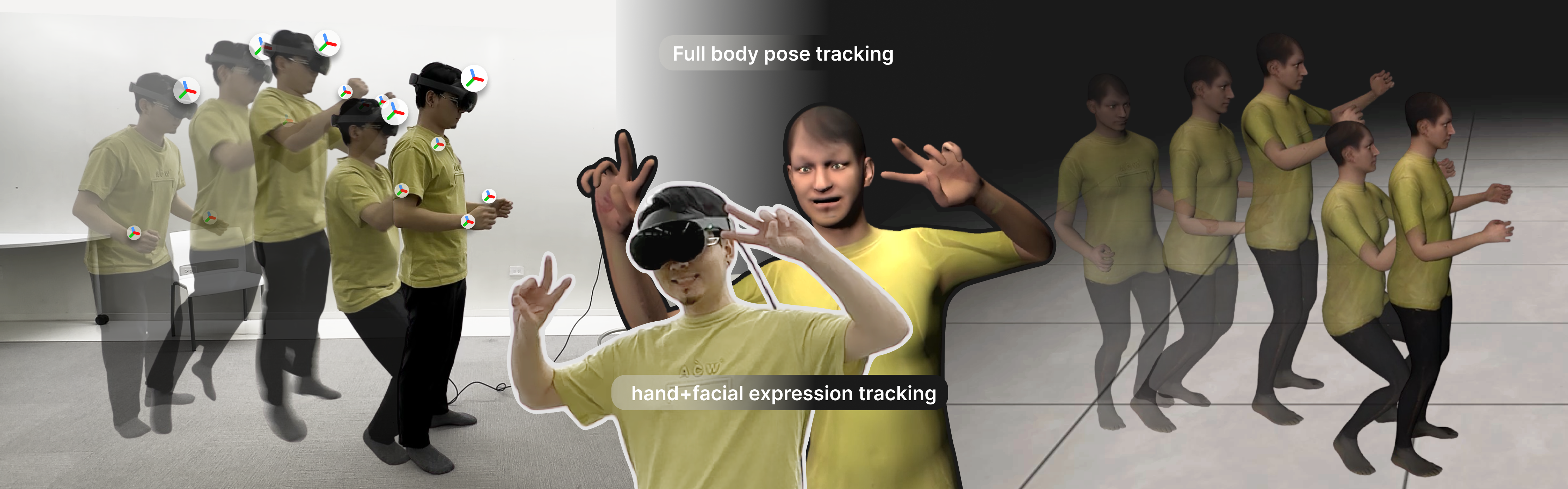}
  \caption{FlowAvatar: a unified system for real-time full-body pose tracking, hand and facial expression tracking, and personalized look-alike appearance on consumer XR devices.}
  \label{fig:teaser}
}

\abstract{
Prior avatar embodiment research has studied tracking completeness \new{(how faithfully the body, hands, and face are sensed and reproduced)} and personalized appearance in isolation, each requiring specialized lab equipment. How these factors interact on consumer hardware remains an open question, because no existing system unifies body pose, face, hands, and personalized appearance on a standalone headset. We present FlowAvatar, a system enabling real-time personalized look-alike avatars with full-body pose, hand articulation, facial expression, and eye gaze tracking from sparse egocentric inputs on consumer XR devices (Meta Quest Pro). Our pose model \new{achieves competitive spatial accuracy and state-of-the-art temporal consistency} on the AMASS benchmark against prior works with comprehensive spatial and temporal losses validated through architecture and loss ablations, while running in real-time. Two user studies (N=12, N=10) provide converging evidence that tracking completeness and personalized appearance serve complementary roles: tracking completeness broadly improves perceived motion quality, while only their combination achieves positive total embodiment.

}

\keywords{User Digitization, Full-Body Pose, XR Avatars}

\begin{document}

\firstsection{Introduction}

\maketitle


\begin{table*}
\caption{Comparison with related work for user digitization on consumer XR devices without additional instrumented sensors.}
\centering
{\fontfamily{\sfdefault}\selectfont
\resizebox{\textwidth}{!}{%
\begin{tabular}{l|cccccccc}
\toprule
Method & Body Pose & Face & Hands & Personalized & On-Device & MPJPE & MPJVE & Jitter \\
\midrule
Meta Quest Full Body~\cite{metaIOBT2023}   & \checkmark & $\times$ & \checkmark & $\times$ & \checkmark & -- & -- & -- \\
Apple Vision Pro Persona~\cite{applevisionpropersonas2024}   & $\times$ & \checkmark & \checkmark & \checkmark & \checkmark & -- & -- & -- \\
Pixel Codec Avatar~\cite{ma2021pixel} & $\times$ & \checkmark & $\times$ & \checkmark & \checkmark & -- & -- & -- \\
AvatarPoser~\cite{jiang2022avatarposer}     & \checkmark & $\times$ & $\times$ & $\times$ & $\times$ & 6.62 & 33.16 & 10.79 \\
AGRoL~\cite{du2023avatars}  & \checkmark & $\times$ & $\times$ & $\times$ & $\times$ & 6.74 & 24.14 & 6.33 \\
AvatarJLM~\cite{zheng2023realistic}  & \checkmark & $\times$ & $\times$ & $\times$ & $\times$ & 5.96 & 27.50 & 6.91 \\
HMDPoser~\cite{dai2024hmd}  & \checkmark & $\times$ & $\times$ & $\times$ & \checkmark & 5.44 & 30.15 & 5.62 \\
\rowcolor{olive!15}
FlowAvatar (our work)       & \checkmark & \checkmark & \checkmark & \checkmark & \checkmark & \textbf{5.40} & \textbf{21.72} & \textbf{3.77} \\
\bottomrule
\end{tabular}}%
}
\label{tab:related-work-comparison}
\end{table*}

In virtual environments, people act through an avatar~\cite{spanlang2014build}, and how strongly they feel that avatar as their own body governs much of the experience. \new{This sense of embodiment rests on three components: body ownership, agency, and self-location~\cite{kilteni2012sense}. When it is strong, users feel more present and extended reality (XR) becomes more effective for social interaction, training, and productivity~\cite{slater2016enhancing}. Unlike immersion, which is the objective sensorimotor fidelity the hardware affords~\cite{slater1997framework}, embodiment is a response of the user.} A growing body of work points to which bodily cues shape it: given a free choice, users building their ideal avatar prioritize faithful control over appearance~\cite{fribourg2020avatar}; adding tracked points such as the feet and pelvis to a head-and-hands avatar raises embodiment~\cite{eubanks2020effects}; and even simple lip-sync of the face strengthens self-identification~\cite{gonzalez2020using}. Together these results point to \emph{tracking completeness}, how many body segments are sensed and faithfully reproduced, as a central driver of embodiment, alongside personalized appearance~\cite{gonzalez2016neurological} and the temporal coherence of motion~\cite{ahuja2021coolmoves,cohn2020snapmove}.

However, each of these studies isolated a single factor and required instrumentation beyond a standalone headset~\cite{fribourg2020avatar,eubanks2020effects,gonzalez2020using}. Two questions remain open. First, how do tracking completeness and personalized appearance each shape embodiment when \emph{combined} in one avatar? Prior work leans toward control over appearance but never tested face and hand completeness alongside personalized texture~\cite{fribourg2020avatar}. Second, do these findings hold when tracking relies on a headset's own sensing, which is sparser and noisier and where temporal artifacts such as jitter especially undermine embodiment~\cite{gonzalez2020self}?

\new{Because these two factors interact rather than act independently, they cannot be teased apart by mixing separate modular systems: swapping components changes several variables at once. Only a single pipeline can hold one factor \emph{fixed} while varying the other.} Yet no consumer system unifies full-body tracking, hands, face, and personalized appearance with sufficient temporal quality (\Cref{tab:related-work-comparison}): Meta lacks face tracking~\cite{metaIOBT2023}, Apple lacks full-body pose~\cite{applevisionpropersonas2024}, and research systems optimize isolated subsystems.

We present FlowAvatar, a unified system that enables real-time personalized look-alike avatars with full-body pose, hands, facial expressions, and eye gaze tracking on off-the-shelf consumer XR devices (\Cref{fig:teaser}). A one-time enrollment generates personalized textures that capture the user's appearance from a brief smartphone video. At runtime, body pose from our neural network, articulated hands and facial expressions from Meta Quest Pro's built-in tracking, and eye gaze all drive a single SMPL-X~\cite{pavlakos2019expressive} mesh simultaneously without additional sensors. Our pose model combines body-part encoding, unified kinematic decoding, and comprehensive spatial and temporal losses, achieving \new{competitive spatial accuracy with state-of-the-art temporal consistency} while running in real time.

Two user studies provide converging evidence on the roles of tracking and appearance in embodiment. In a motion quality study (N=12), adding face and hand tracking significantly improved all six perceived quality dimensions, while adding personalized texture reached significance only for Interaction Readiness. In an embodiment study (N=10), tracking completeness was associated with \emph{agency}, while personalized appearance was associated with \emph{ownership}, and only the condition combining both achieved positive total embodiment. These results suggest that tracking completeness and personalized appearance serve complementary roles, and that platform designers aiming for full embodiment should invest in both rather than treating them as competing priorities.

This paper contributes:
\begin{enumerate}
\setlength{\itemsep}{0pt}
\setlength{\parskip}{0pt}
\setlength{\parsep}{0pt}
\item \new{\textbf{FlowAvatar}, the first open-source avatar system to holistically integrate real-time full-body pose, hand articulation, facial expression, and eye gaze tracking with personalized appearance on a standalone consumer XR device, enabling the study and evaluation of a complete avatar under realistic consumer constraints.}
\item \new{A \textbf{full-body pose framework} that infers pose from sparse egocentric head and hand inputs through body-part encoding, unified kinematic decoding, and comprehensive spatial and temporal losses, achieving competitive spatial accuracy and state-of-the-art temporal consistency on the AMASS dataset, and released as open source.}
\item \new{\textbf{Two user studies}, enabled by this integrated pipeline, that disentangle the complementary roles of tracking completeness and personalized appearance in shaping perceived motion quality and embodiment.}
\end{enumerate}

\section{Related Work}

Our work sits at the intersection of graphics, perception, and extended reality body capture. We are not the first to recognize the value of digitizing user bodies, capturing personal appearance, and creating embodied avatars. Thus, we first succinctly review related work in motion capture, sensed both externally and by worn hardware. We then discuss research efforts focused towards user digitization from XR headsets, and their effects on embodiment and avatar representation.

\subsection{User Digitization from External Sensors}

There has been significant progress in generating digital representations using external sensors and capture rigs. Commercial motion capture systems such as Vicon use specialized hardware, worn retroreflective markers, and multi-camera rigs to create millimeter-accurate digitizations. Methods such as Holoportation~\cite{orts2016holoportation} and Starline~\cite{russell2021project} employ a suite of external sensors (RGB, depth, IR) to generate photorealistic representations of humans and environment. While compelling, their prohibitive costs make them impractical for everyday use.

Therefore, much research has been devoted to instrumentation-free approaches that rely on computer vision techniques. Numerous works have looked into particular verticals such as tracking eyes, hands, or the face~\cite{baltruvsaitis2016openface}. For unified full-body tracking, works have predominantly made use of a combination of depth and RGB~\cite{goel2023humans} cameras.

Photorealistic avatar creation from RGB has progressed from mesh-based texture reconstruction~\cite{zhi2020texmesh} to neural rendering methods that generate animatable avatars from monocular video~\cite{kerbl20233d,hu2024gaussianavatar}, with multi-camera setups further enhancing quality and drivability~\cite{saito2024relightable}. Recent work has also begun distilling these high-fidelity Gaussian avatars into mobile-ready representations, enabling real-time rendering of multiple avatars on standalone VR headsets~\cite{iandola2025squeezeme}.

\subsection{User Digitization from Body Worn Sensors}

Wearable sensors provide a portable alternative to external sensors. Since wearables can be intrusive, research has often focused on digitizing specific body parts. Hands, for example, have been digitized through wrist-worn cameras~\cite{wu2020back}, EMG~\cite{liu2021neuropose}, EIT~\cite{kyu2024eitpose}, antennas~\cite{kim2022etherpose}, and depth armbands~\cite{devrio2022discoband}. For full-body pose, body-worn IMUs are most common. Commercial systems such as Xsens use many high-grade inertial sensors (typically 17), and later work has reduced this count to 6~\cite{yi2021transpose} and even 3~\cite{van2024diffusionposer}. More recently, work has explored using IMUs found in everyday consumer devices~\cite{xu2024mobileposer,mollyn2023imuposer}, albeit producing a lower-fidelity full-body pose estimate.

\subsection{User Digitization from XR Headsets}

Many works instrument either the headset~\cite{ahuja2019mecap, chen2024egoavatar}, or the wearer~\cite{dai2024hmd} with additional sensors to digitize parts of their body in XR. Although these approaches can achieve high fidelity, they typically require significant server-side GPU compute, increase system complexity, and reduce user comfort. This has motivated a shift towards instrumentation-free user digitization using only the sensing capabilities of standard headsets.

Recently, instrumentation-free methods have advanced specific digitization components: photorealistic face capture~\cite{saito2024relightable} and mesh-based body reconstruction~\cite{aliakbarian2023hmd}. The works most related to ours estimate full-body pose from an XR headset’s sparse observations (head + hands). Approaches such as AvatarPoser~\cite{jiang2022avatarposer}, HMDPoser~\cite{dai2024hmd}, and AGRoL~\cite{du2023avatars} use data-driven models with SMPL~\cite{bogo2016smpl}, often assuming continuous hand visibility in the headset’s FoV. To relax this assumption, EgoPoser~\cite{jiang2024egoposer} incorporates random hand masking or FoV modeling, but typically relies on spatio-temporal transformers with substantial computational cost, which complicates real-time on-device deployment.

Across these methods, a trend toward increasing architectural complexity is evident: from RNNs~\cite{jiang2022avatarposer} to spatio-temporal Transformers~\cite{jiang2024egoposer} to conditional diffusion models~\cite{du2023avatars}. Besides architecture choice, these methods also vary in their use of temporal supervision: HMDPoser~\cite{dai2024hmd} employs an acceleration loss, AvatarJLM~\cite{zheng2023realistic} uses multi-scale velocity losses at multiple frame strides, while AvatarPoser~\cite{jiang2022avatarposer}, AGRoL~\cite{du2023avatars}, and EgoPoser~\cite{jiang2024egoposer} rely solely on spatial objectives. Our work combines both acceleration and velocity supervision with a lightweight recurrent architecture, and we ablate both dimensions independently in \Cref{subsec:ablation}. FlowAvatar builds on this body of work as a unified system connecting personalized appearance generation to real-time body, hand, face, and eye tracking on consumer XR hardware.


\subsection{Embodiment and Avatar Representation}
\label{sec:embodiment-related}

Avatars are the primary channel for self-representation in XR. Higher-fidelity avatars increase trust and engagement from third-person perspectives~\cite{izzouzi2024influence} and strengthen first-person embodiment~\cite{peck2021avatar}, which has been linked to improved empathy~\cite{peck2013putting}, presence~\cite{slater2010simulating}, and cognitive performance~\cite{steed2016impact}. \new{The single-factor studies noted earlier~\cite{fribourg2020avatar,eubanks2020effects,gonzalez2020using} exemplify a broader literature that a recent meta-analysis of body ownership illusions synthesizes: both tracking fidelity and avatar appearance reliably shape embodiment, yet their relative contributions remain hard to compare because studies manipulate them in isolation and under heterogeneous conditions~\cite{mottelson2023systematic}.}

Beyond these findings, Debarba et al.~\cite{debarba2022plausibility} showed that hands and face contribute disproportionately to perceived animation plausibility compared to gross body segments. On the appearance side, look-alike avatars elicit greater neural self-recognition than generic ones~\cite{gonzalez2016neurological}, with personalized avatars increasing self-identification over cartoon representations~\cite{salagean2023meeting}. Tao et al.~\cite{tao2023embodying} found physics-aware avatar motion can improve embodiment even with small motion discrepancies. \new{Even recent work examines these factors one at a time: Tschanter et al.~\cite{tschanter2026usecase} find that the value of embodied tracking depends on the task, while Fiedler et al.~\cite{fiedler2026ami} show that avatar appearance and its alignment with the user's real body influence self-perception and behavior. Personalized look-alike avatars are meanwhile becoming widely accessible through smartphone-based reconstruction, removing the need for capture rigs~\cite{menzel2025avatars}.}

FlowAvatar brings these separately studied components into one driven avatar on consumer hardware, letting us study tracking completeness and personalized appearance together rather than one at a time, a comparison prior work could not make (\Cref{sec:user-studies}).

\section{Pose Estimation Pipeline}
\label{sec:pose-estimation}

\new{FlowAvatar builds and drives a personalized avatar in two stages: a one-time enrollment that generates the look-alike avatar from a short phone video, and a real-time runtime that animates it on the headset. We present the runtime pose estimation first, as the technical core, and cover enrollment and deployment in \Cref{sec:avatar-pipeline}.} The avatar is represented using the SMPL-X parametric body model~\cite{pavlakos2019expressive}, a single mesh controlled by body shape ($\beta$) for proportions, body pose ($\theta$) for joint rotations, hand articulation, and facial blendshapes for expressions. This section describes how we animate that mesh in real time from sparse headset signals: a neural network infers full-body pose from head and hand tracking (\Cref{subsec:input-rep}--\Cref{subsec:losses}), and the estimated body pose is then combined with hand, face, and eye tracking from the headset's built-in sensors (\Cref{subsec:face-hand}).

\subsection{Body Pose Estimation}

\subsubsection{Input and Output Representation}
\label{subsec:input-rep}

Following prior work~\cite{dai2024hmd,jiang2022avatarposer}, we represent each tracked component with position ($p \in \mathbb{R}^3$) in the global coordinate system, orientation ($r \in \mathbb{R}^6$) using the 6D rotation representation, angular velocity ($\omega \in \mathbb{R}^6$) and linear velocity ($v \in \mathbb{R}^3$) computed via finite differences.

For each frame $t$, we construct input vectors for the head ($x_h^t \in \mathbb{R}^{18}$), left hand ($x_{lh}^t \in \mathbb{R}^{18}$), and right hand ($x_{rh}^t \in \mathbb{R}^{18}$), where each vector concatenates position, orientation, angular velocity, and linear velocity. Additionally, we compute head-relative representations of the hand positions ($x_{lh/h}^t, x_{rh/h}^t \in \mathbb{R}^{18}$), which provide valuable cues for estimating body proportions and improving joint position accuracy. The complete input at time $t$ is:
\begin{equation}
    x^t = [x_h^t, x_{lh}^t, x_{rh}^t, x_{lh/h}^t, x_{rh/h}^t] \in \mathbb{R}^{90}
\end{equation}

To capture temporal dynamics, we process sequences of $T=40$ consecutive frames during both training and inference, corresponding to $\sim$667~ms at our device capture frequency of 60~Hz. During inference, we employ a sliding window approach where each prediction uses the current frame and the previous $T-1$ frames, enabling real-time performance while maintaining temporal consistency. The system outputs SMPL-X~\cite{pavlakos2019expressive} body parameters including joint rotations ($\theta_{pose}^t \in \mathbb{R}^{6 \times 22}$) and shape parameters ($\beta \in \mathbb{R}^{10}$), which are processed through forward kinematics to drive the final 3D mesh with articulated pose.

\subsubsection{Architecture}
\label{subsec:architecture}

\begin{figure}[t]
  \centering
  \includegraphics[width=1\linewidth]{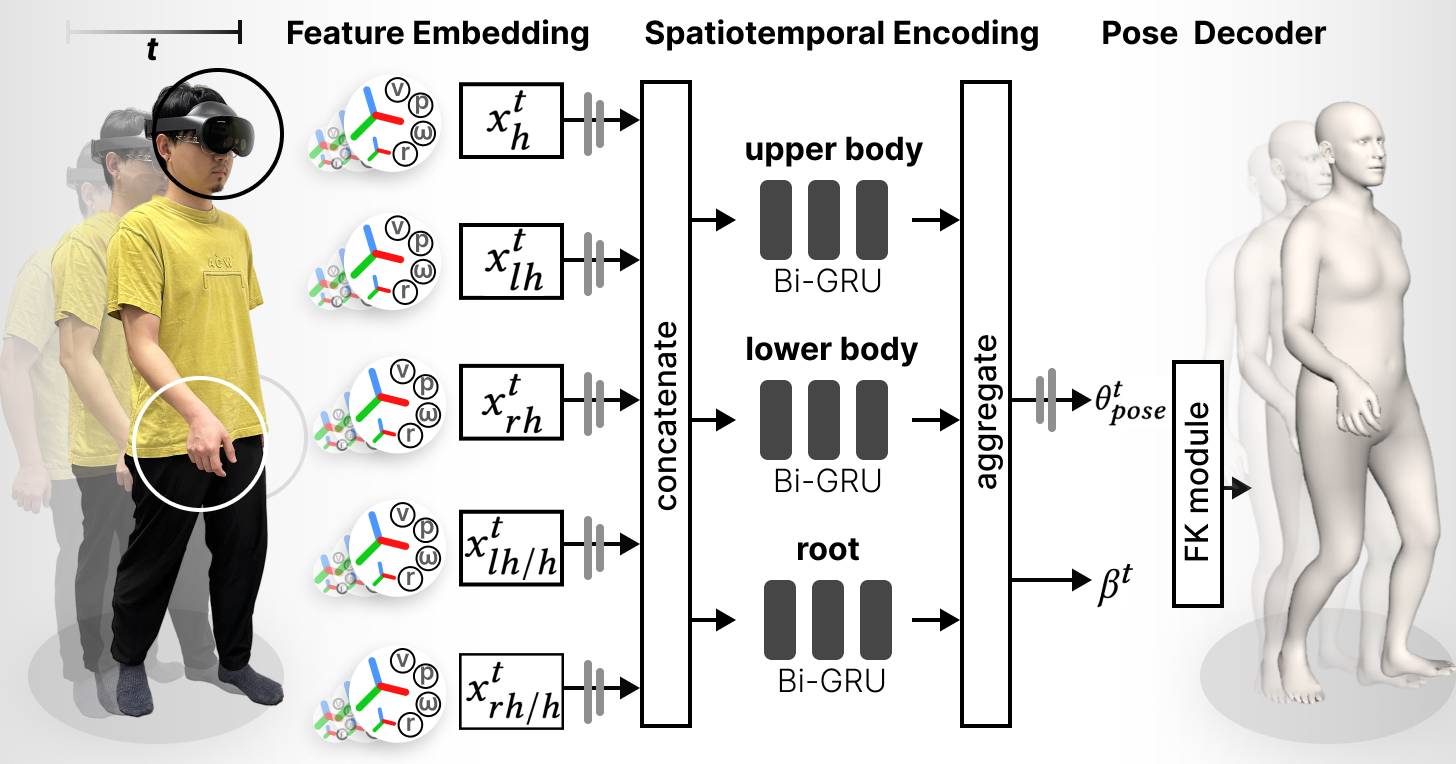}
  \caption{Pose estimation architecture: three parallel BiGRU networks (upper body, lower body, root) process shared embedded features to generate SMPL-X parameters.}
  \label{fig:pose_pipeline}
\end{figure}

For inputs $x^t \in \mathbb{R}^{90}$ at time $t$, we first apply a feature embedding module to transform these sparse tracking signals into higher-dimensional representations $f_e^t = \phi_e(x^t) \in \mathbb{R}^{d_e}$, where $\phi_e$ is implemented as a two-layer MLP with ReLU activations and $d_e = 256$. We concatenate the raw inputs before embedding for parameter efficiency.

After feature embedding, we process the embedded features through three parallel bidirectional GRU networks, each specialized for a different body region (upper body, lower body, and root). All three GRUs receive the same embedded input features but learn region-specific temporal dynamics:
\begin{equation}
h^{part}_t = \text{BiGRU}_{part}(f_e^t, h^{part}_{t-1})
\end{equation}
where each $h^{part}_t$ represents the hidden state for a specific body region at time $t$ and $d_h = 512$. Together, all three GRUs model the spatio-temporal patterns across all body regions. The outputs are then fed into a unified decoder to generate the SMPL-X body parameters:
\begin{equation}
\theta_{pose}^t = \phi_{pose}(h_t)
\end{equation}
The decoder $\phi_{pose}$ consists of a two-layer MLP with ReLU activations and dropout ($p=0.1$) that outputs the complete pose parameters $\theta_{pose}^t \in \mathbb{R}^{6 \times 22}$ for all body joints at time $t$. Along with the predicted shape parameters $\beta^t \in \mathbb{R}^{10}$, these are processed through SMPL-X forward kinematics to generate the final 3D joint positions $P^t \in \mathbb{R}^{22 \times 3}$ and the corresponding mesh with articulated pose. We select GRU over Transformer as it achieves better accuracy across all metrics under identical training configurations (\Cref{subsec:ablation}).

\subsubsection{Training Objectives}
\label{subsec:losses}

We employ multiple complementary loss functions organized into four categories to guide our model toward accurate, temporally consistent motion.

\textbf{Pose loss.} We employ L1 root orientation loss ($\mathcal{L}_{ori}$), L1 local joint rotation loss ($\mathcal{L}_{lrot}$), L1 global joint rotation loss ($\mathcal{L}_{grot}$), and L1 joint position loss ($\mathcal{L}_{pos}$), following previous approaches~\cite{dai2024hmd,jiang2024egoposer}. The local rotation loss ensures accurate joint angles, while the global rotation loss helps reduce error accumulation along the kinematic chain. Joint position loss provides direct supervision on the spatial configuration of the body.

\textbf{Temporal loss.} For temporal consistency, we adopt an acceleration loss that penalizes deviations in joint accelerations between predicted and ground-truth motions:
\begin{equation}
    \mathcal{L}_{acc} = \frac{1}{(T-2) \cdot J}\sum_{t=1}^{T-2}\sum_{j=1}^{J}\|a_j^t - \hat{a}_j^t\|_1
\end{equation}
where $a_j^t$ is the acceleration of joint $j$ at time $t$, computed via second-order finite differences. Additionally, we implement a multi-scale velocity loss that measures velocity differences at varying time scales:
\begin{equation}
    \mathcal{L}_v(l) = \frac{1}{(T-l) \cdot J}\sum_{t=1}^{T-l}\sum_{j=1}^{J}\|(P_j^{t+l} - P_j^t) - (\hat{P}_j^{t+l} - \hat{P}_j^t)\|_1
\end{equation}

\textbf{Hand alignment loss.} Standard inverse kinematics post-processing~\cite{jiang2022avatarposer} iteratively adjusts predicted joint angles to match observed hand positions at inference time, adding computational overhead and introducing discontinuities between corrected and uncorrected frames. We instead integrate a learned hand alignment loss that maintains end-to-end trainability:
\begin{equation}
    \mathcal{L}_{h} = \frac{1}{2T}\sum_{t=1}^{T}(\|P_{lh}^t - \hat{P}_{lh}^t\|_1 + \|P_{rh}^t - \hat{P}_{rh}^t\|_1)
\end{equation}
where $P_{lh}^t$ and $P_{rh}^t$ are the predicted positions of left and right hands, ensuring accurate hand placement without runtime post-processing.

\textbf{Ground contact loss.} We employ two complementary losses for ground contact regularization. The foot contact loss enforces zero velocity when feet are in contact with the ground:
\begin{equation}
    \mathcal{L}_{fc} = \frac{1}{T-1}\sum_{t=1}^{T-1}\|(P_{feet}^{t+1} - P_{feet}^t) \cdot m^t\|_1
\end{equation}
where $m^t \in \{0,1\}^{n_{feet}}$ is a binary foot contact mask, preventing the common ``skating'' artifact. We also implement a ground penetration loss:
\begin{equation}
    \mathcal{L}_{p} = \frac{1}{T}\sum_{t=1}^{T}\|(z_{min}^t - z_{ground}) \cdot l^t\|_1
\end{equation}
where $z_{ground}$ is the ground height, $z_{min}^t$ is the height of the lowest joint, and $l^t \in \{0,1\}^{J}$ indicates whether a joint is below ground. Together, these constraints address the two most common artifacts in egocentric pose estimation, foot skating and ground penetration, which we evaluate quantitatively in~\Cref{subsec:ablation}.

Our complete loss function is defined as:
\begin{equation}
\begin{split}
    \mathcal{L} = &\lambda_{ori}\mathcal{L}_{ori} + \lambda_{lrot}\mathcal{L}_{lrot} + \lambda_{grot}\mathcal{L}_{grot} + \lambda_{pos}\mathcal{L}_{pos} + \\
    &\lambda_{acc}\mathcal{L}_{acc} + \lambda_{vel}\mathcal{L}_{vel} + \lambda_{h}\mathcal{L}_{h} + \\
    &\lambda_{fc}\mathcal{L}_{fc} + \lambda_{p}\mathcal{L}_{p}
\end{split}
\label{eq:loss}
\end{equation}

\subsection{Hand, Face, and Eye Pose Integration}
\label{subsec:face-hand}

\begin{figure}[t]
  \centering
  \includegraphics[width=1\linewidth]{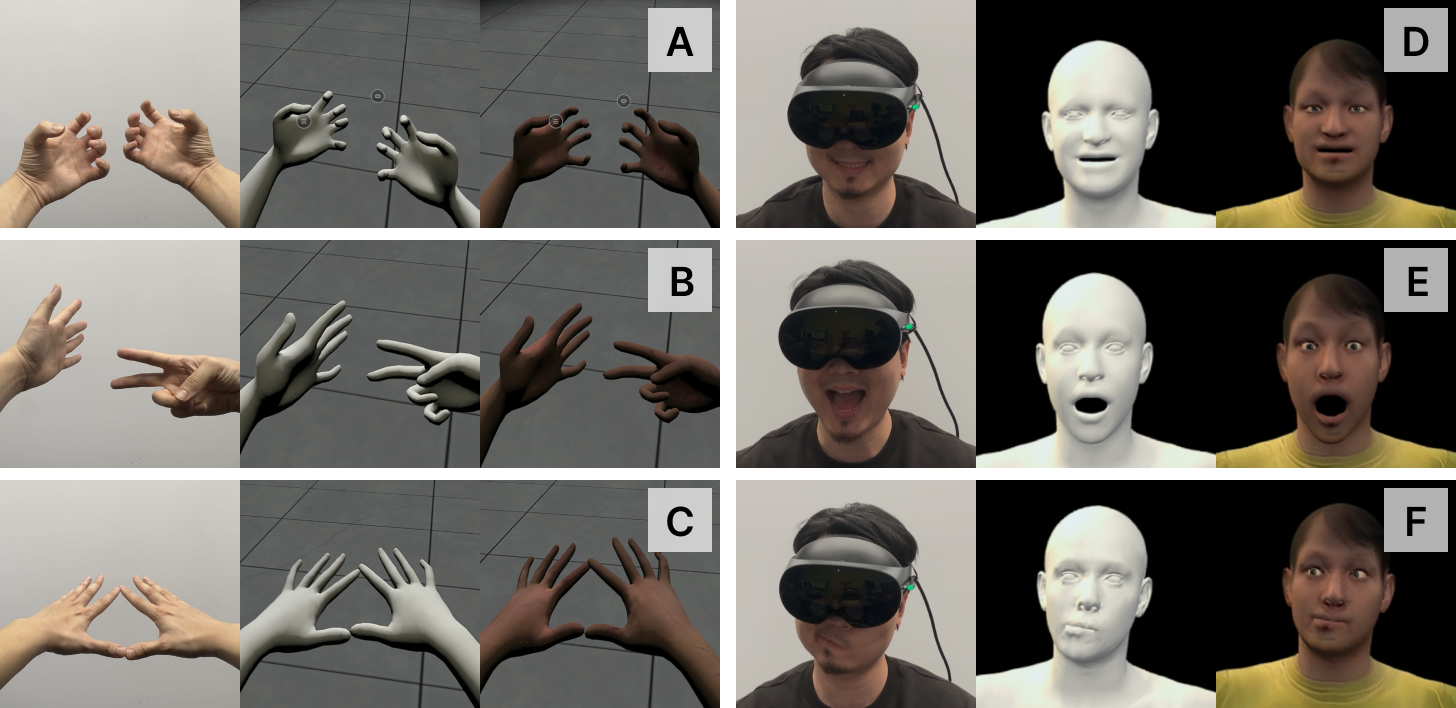}
  \caption{Examples of hand (A--C) and facial expression (D--F) capture on Meta Quest, remapped to SMPL-X mesh.}
  \label{fig:face_hand_integration}
\end{figure}

Beyond body pose, the Quest Pro's built-in sensors provide hand articulation, facial expressions, and eye gaze. Each of these operates in a different parameter space from SMPL-X, so integrating them onto the same mesh requires mapping for each modality (\Cref{fig:face_hand_integration}).

For hand tracking, we perform a one-time offline rig setup in Autodesk Maya, retargeting Meta's 18-joint OVR hand skeleton (with an extra joint on the thumb and pinky) onto SMPL-X's 16-joint hand (15 finger joints plus wrist) to produce a rigged SMPL-X mesh that we then import into our Unity runtime. At runtime, the retargeted hand poses blend with the body pose from our neural network, producing coordinated full-body motion with articulated fingers.

For facial expressions, at runtime we map Meta's 52 ARKit-compatible blendshape coefficients~\cite{appleARKit} to SMPL-X's expression space using a conversion matrix from EMAGE~\cite{liu2024emage}, giving 100 expression parameters plus 3 jaw rotations. Because our Unity runtime exposes only 10 expression blendshapes, we reduce these 100 coefficients to those 10 through an empirically calibrated projection, and pass the jaw rotations through directly. This trades subtle detail for real-time rendering while preserving the most salient movements such as smiling, frowning, mouth opening, and brow movement.

Additionally, we extract eye movement data from Meta's gaze tracking to control SMPL-X eyeball rotations (yaw and pitch for each eye), enabling natural eye contact in social scenarios.

\section{Full-Body Avatar Pipeline}
\label{sec:avatar-pipeline}

With the pose estimation pipeline in place, we now describe how FlowAvatar produces and delivers the complete look-alike avatar in two stages: a one-time enrollment that builds a personalized avatar offline from a brief smartphone video (\Cref{subsec:texture-gen}), and a runtime that drives it on the headset (\Cref{sec:deployment}). \new{We build on established differentiable rendering~\cite{Svitov_2024_ACCV} and enroll each user from a single phone video, the standard, low-friction workflow that commercial systems such as Meta Horizon~\cite{metaAvatars} use to keep avatar creation accessible.}


\subsection{Enrollment and Texture Generation}
\label{subsec:texture-gen}
\label{sec:system-overview}

The user opens a companion smartphone app and records a $\sim$10-second video while slowly completing a full turn using the front-facing camera. No special lighting, background, or calibration is required. From this video, our pipeline generates a personalized UV texture map and estimates initial body-shape ($\beta$) parameters through the following four stages.

\begin{figure}[t]
  \centering
  \includegraphics[width=1\linewidth]{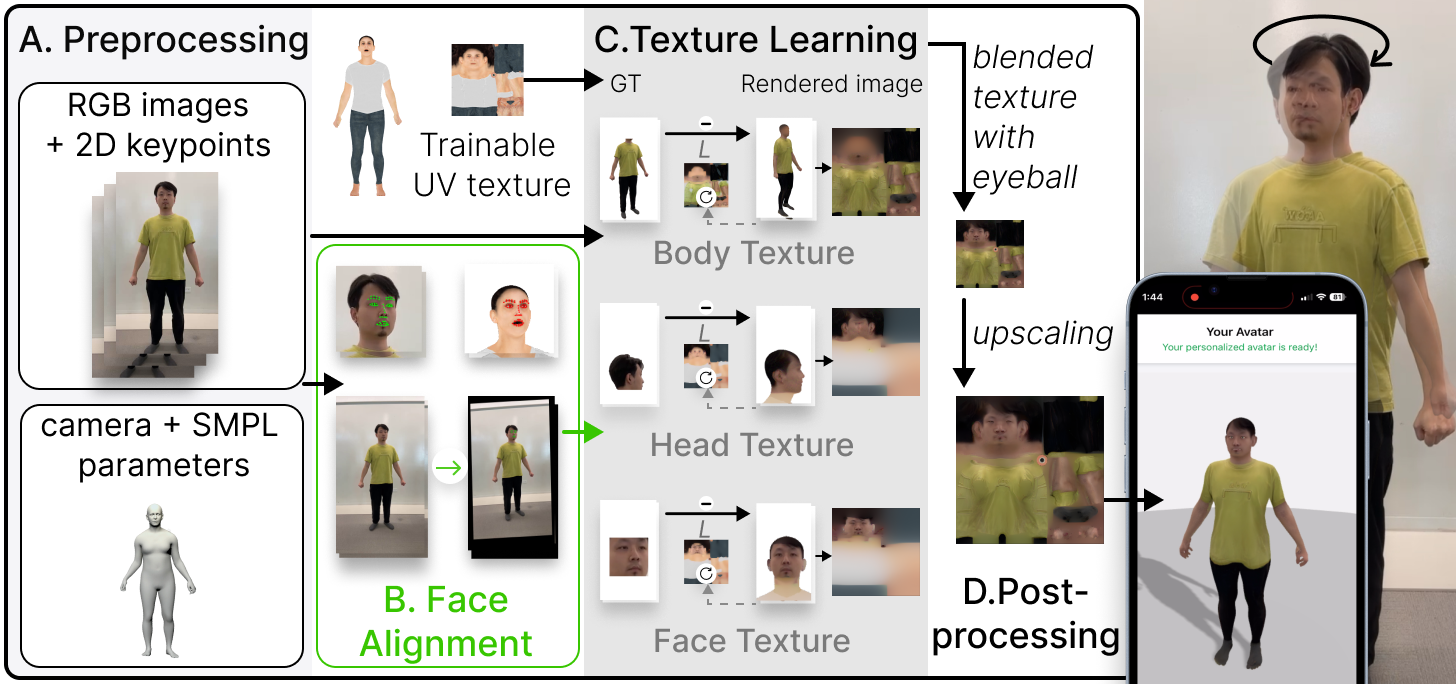}
  \caption{Texture generation pipeline: (A) preprocessing, (B) face alignment, (C) region-specific differentiable rendering, (D) compositing and upscaling.}
  \label{fig:texture_pipeline}
\end{figure}

\textbf{A. Preprocessing.} We estimate SMPL-X shape, pose, and camera parameters for each frame ($\sim$250 frames at 25~FPS) using PyMAF-X~\cite{zhang2023pymaf}, and segment body regions from 2D pose~\cite{mmpose2020} and a human segmentation mask~\cite{chen2022robust}. To mitigate inaccuracies in camera parameter estimation, we \new{adopt a three-component texture design, training} separate textures for the body, face, and head using corresponding region masks\new{, so each region can be aligned and optimized independently}.

\textbf{B. Face alignment.} PyMAF-X assumes a weak-perspective camera model, which causes misalignment between rendered and input images, particularly degrading face texture quality under perspective distortion. We correct this with a 2D affine transform computed from facial landmarks~\cite{Zhou_2023_CVPR} detected in both rendered and input images, aligning the input to serve as ground truth for face texture training.

\textbf{C. Texture learning.} Using the estimated SMPL-X and camera parameters, we render 2D images with trainable UV textures via a differentiable renderer~\cite{Laine2020diffrast} and optimize against the input RGB frames (\Cref{fig:texture_pipeline}). Following Svitov et al.~\cite{Svitov_2024_ACCV}, the texture loss combines $L_2$ reconstruction, LPIPS~\cite{zhang2018perceptual}, SSIM, and total variation regularization. Textures for the face, head, and body are trained separately with corresponding masks, typically converging within 500 iterations.

\textbf{D. Postprocessing.} The learned face, head, and body textures are blended into a single $512\times512$ texture, with an eyeball texture from the reference template composited into the final map. The result is upscaled to $2048\times2048$ via Real-ESRGAN~\cite{wang2021real} to reduce noise and improve detail. Shape betas from the texture generation process are applied to the SMPL-X mesh at runtime to customize body proportions. Total processing time is approximately 15 minutes on an NVIDIA RTX 4090. The user can preview the resulting avatar on the phone before proceeding to XR.

\subsection{Runtime and Deployment}
\label{sec:deployment}

The user puts on a Quest Pro and logs in to fetch the avatar texture generated during enrollment (\Cref{subsec:texture-gen}). No calibration or T-pose is needed. At each frame, the pose estimation model infers full-body pose from the headset's head and hand tracking (\Cref{sec:pose-estimation}), while the Quest Pro's built-in sensors provide hand articulation, facial expressions, and eye gaze. These are combined onto the same SMPL-X mesh (\Cref{subsec:face-hand}), producing a coherent avatar with synchronized body, hands, face, and eyes (\Cref{fig:teaser}). The model also predicts shape ($\beta$) parameters, which are used once at startup to refine the body proportions estimated during enrollment.

FlowAvatar supports two deployment configurations, both of which we release as open source at \url{https://github.com/SPICExLAB/FlowAvatar}: a streaming configuration and an on-device configuration.

In the \textbf{streaming configuration}, Meta Quest Pro tracking data is streamed to a Windows laptop with an RTX 4090 for inference. The end-to-end pipeline runs at 247~FPS under this setup; HMDPoser~\cite{dai2024hmd} achieves 43.3~FPS under the same configuration. The pose model alone runs at 91.5~FPS on a general-purpose CPU (Apple M1 Max), increasing to 341.0~FPS after ONNX optimization (3.7$\times$). All user studies reported in this paper use the streaming configuration.

In the \textbf{on-device configuration}, FlowAvatar runs fully on Meta Quest Pro (CPU) via Unity Sentis's ONNX runtime. \new{Because Unity Sentis lacks GRU operator support, the on-device model uses the LSTM variation, a modest accuracy drop (MPJRE 4.43 vs.\ the GRU's 4.19; \Cref{tab:ablation_backbone}).} We sample head, hand, and face into a circular buffer and run capture and inference asynchronously. The complete pipeline sustains 27~FPS at a 97~ms end-to-end \new{motion delay}, enabling practical untethered use. \new{The hand ($\sim$50~ms) and face ($\sim$70~ms) trackers run asynchronously on native Meta threads and the pose loop reads their latest state, so the slowest tracker sets the delay but not the frame rate. This motion delay stays well within the $\sim$130--170~ms that prior work finds tolerable for avatar motion~\cite{halbhuber2023understanding}. We have also deployed the GRU model on AndroidXR (Samsung Moohan, Snapdragon XR2 Gen~2) via TFLite at 60~Hz with no accuracy loss.}

\section{Evaluation}
\label{sec:evaluation}


Before examining how tracking completeness and personalized appearance affect embodiment (\Cref{sec:user-studies}), this section validates the pose estimation model through benchmark comparison, ablation studies, and input robustness (\Cref{subsec:pose-eval}), followed by a preliminary texture generation evaluation (\Cref{subsec:texture-eval}).

\subsection{Full-body Pose Estimation}
\label{subsec:pose-eval}


Following recent common practice~\cite{du2023avatars,dai2024hmd,jiang2022avatarposer,zheng2023realistic}, we evaluate on the public AMASS dataset~\cite{mahmood2019amass} using the train/test split introduced by HMDPoser~\cite{dai2024hmd}: twelve AMASS subsets for training with HumanEva~\cite{sigal2010humaneva} and Transition held out for testing, and ground-truth body shape parameters used to compute joint positions. This split tests generalization to entirely unseen motion datasets rather than random partitions within the same subsets. We train with Adam optimizer (initial learning rate $10^{-4}$ decaying to $10^{-5}$ after 75\% of training), batch size 256, for 400 epochs on an RTX 4090. For our loss (\Cref{eq:loss}), we set $\lambda_{ori} = 1.0$, $\lambda_{lrot} = 5.0$, $\lambda_{grot} = 1.0$, $\lambda_{pos} = 5.0$, $\lambda_{acc} = 0.5$, $\lambda_{vel} = 5.0$, $\lambda_{h} = 5.0$, $\lambda_{fc} = 1.0$, and $\lambda_{p} = 1.0$.

All benchmark results below use the streaming configuration described in \Cref{sec:deployment} with the 3-layer BiGRU model. Following established practices~\cite{du2023avatars,dai2024hmd}, we report spatial metrics, including MPJRE (mean joint rotation error in degrees) and MPJPE (mean joint position error in cm), alongside temporal metrics, including MPJVE (mean joint velocity error in cm/s) and Jitter (third derivative of position, $10^{2}$~m/s$^3$), which reflects the smoothness of generated motion.

\subsubsection{Comparison with Prior Work}
\label{subsec:benchmark-results}


\begin{table}[b]
\caption{FlowAvatar compared with prior works on AMASS.}
  \centering
  \begingroup
  \renewcommand{\arraystretch}{1.05}
  \begin{tabularx}{\columnwidth}{l|*{4}{>{\centering\arraybackslash}X}}
    \toprule
    Method & MPJRE$\downarrow$ & MPJPE$\downarrow$ & MPJVE$\downarrow$ & Jit.$\downarrow$ \\
    \midrule
    AvatarPoser~\cite{jiang2022avatarposer} & 4.68 & 6.62 & 33.16 & 10.79 \\
    AGRoL~\cite{du2023avatars}              & 4.38 & 6.74 & 24.14 & 6.33 \\
    AvatarJLM~\cite{zheng2023realistic}     & 4.45 & 5.96 & 27.50 & 6.91 \\
    HMDPoser~\cite{dai2024hmd}              & 4.27 & 5.44 & 30.15 & 5.62 \\
    EgoPoser~\cite{jiang2024egoposer}       & 4.45 & 6.85 & 35.20 & 18.64 \\
    \rowcolor{olive!15}
    FlowAvatar (ours)                       & \textbf{4.19} & \textbf{5.40} & \textbf{21.72} & \textbf{3.77} \\
    \bottomrule
  \end{tabularx}
  \endgroup
  \label{tab:comparison}
\end{table}

\Cref{tab:comparison} shows that FlowAvatar achieves competitive spatial accuracy and state-of-the-art temporal consistency on the AMASS benchmark. Compared to HMDPoser~\cite{dai2024hmd}, FlowAvatar matches its spatial accuracy (MPJPE: 5.40~cm vs 5.44~cm) while substantially improving temporal metrics: MPJVE of 21.72~cm/s and jitter of 3.77~$10^{2}$~m/s$^3$ represent 28\% and 33\% improvements respectively over HMDPoser's 30.15~cm/s and 5.62~$10^{2}$~m/s$^3$. AvatarPoser~\cite{jiang2022avatarposer} shows high jitter (10.79~$10^{2}$~m/s$^3$) and MPJVE (33.16~cm/s), and AvatarJLM~\cite{zheng2023realistic} likewise has elevated jitter (6.91~$10^{2}$~m/s$^3$) and velocity error (27.50~cm/s). AGRoL~\cite{du2023avatars} reaches the closest temporal metrics among the baselines but higher positional error (MPJPE: 6.74~cm). \new{EgoPoser~\cite{jiang2024egoposer} shows the highest velocity error (35.20~cm/s) and jitter (18.64~$10^{2}$~m/s$^3$, nearly $5\times$ ours), reflecting a focus on large-space global positioning over temporal smoothness.} Since spatial offsets below 10 cm are typically imperceptible~\cite{cohn2020snapmove}, all methods fall within the perceptual threshold on position, making temporal artifacts the primary differentiator.

\subsubsection{Ablation Studies}
\label{subsec:ablation}

\textbf{Loss ablation.} \Cref{tab:ablation_loss} shows the impact of each loss component. Removing temporal losses increases MPJVE from 21.72 to 25.58~cm/s and jitter from 3.77 to 5.17~$10^{2}$~m/s$^3$, representing 18\% and 37\% degradation respectively, the largest single-category impact. Without hand alignment loss, H-PE increases from 2.31~cm to 2.98~cm (29\% worse). The ground contact losses show trade-offs: their removal slightly improves hand accuracy (2.25~cm vs 2.31~cm) but degrades jitter (4.05 vs 3.77~$10^{2}$~m/s$^3$) and leads to visually noticeable foot-floating artifacts during high-intensity motions. The combination of all losses achieves the best overall balance.

\begin{table}[h]
\caption{Ablation across different loss configurations.}
  \centering
  \begingroup
  \renewcommand{\arraystretch}{1.05}
  \begin{tabularx}{\columnwidth}{l|*{4}{>{\centering\arraybackslash}X}}
    \toprule
    Strategy & MPJRE$\downarrow$ & MPJVE$\downarrow$ & Jitter$\downarrow$ & H-PE$\downarrow$ \\
    \midrule
    No temporal loss & 4.55 & 25.58 & 5.17 & 2.50 \\
    No hand loss & 4.41 & 23.26 & 3.91 & 2.98 \\
    No ground loss & 4.31 & 22.55 & 4.05 & \textbf{2.25} \\
    \rowcolor{olive!15}
    All losses & \textbf{4.19} & \textbf{21.72} & \textbf{3.77} & 2.31 \\
    \bottomrule
  \end{tabularx}
  \endgroup
  \label{tab:ablation_loss}
\end{table}



\textbf{Backbone comparison.} \Cref{tab:ablation_backbone} compares four backbone architectures under identical training configurations, with FPS measured on an Apple M1 Max CPU for model-only inference. Each variant uses the same feature embedding and decoder described in \Cref{subsec:architecture}, swapping only the backbone that processes temporal sequences. GRU achieves the best accuracy across all metrics at the lowest computational cost (154M FLOPs, 565 FPS). LSTM provides the closest accuracy with comparable speed, making it a viable alternative when GRU operators are unavailable (\Cref{sec:deployment}). The Transformer and FastViT~\cite{vasu2023fastvit} lag behind on all accuracy metrics despite higher computational cost.

\begin{table}[h]
\caption{Backbone architecture comparison (all 3-layer).}
  \centering
  \begingroup
  \setlength{\tabcolsep}{3pt}
  \renewcommand{\arraystretch}{1.05}
  \begin{tabularx}{\columnwidth}{l|*{5}{>{\centering\arraybackslash}X}}
    \toprule
    Backbone & FLOPs & FPS & MPJRE$\downarrow$ & MPJPE$\downarrow$ & MPJVE$\downarrow$ \\
    \midrule
    Transformer & 171M & 308 & 4.83 & 6.94 & 29.55 \\
    FastViT~\cite{vasu2023fastvit} & 1027M & 39.1 & 5.49 & 7.65 & 29.01 \\
    LSTM & 190M & 542.3 & 4.43 & 5.75 & 22.95 \\
    \rowcolor{olive!15}
    GRU (ours) & 154M & 565.0 & \textbf{4.19} & \textbf{5.40} & \textbf{21.72} \\
    \bottomrule
  \end{tabularx}
  \endgroup
  \label{tab:ablation_backbone}
\end{table}


\textbf{Input robustness.} FlowAvatar stays usable under severe tracking loss: with only the head and a single hand available at test time (the model trained with 40\% hand-masking augmentation), MPJRE rises modestly from 4.19° to 5.29°.

\subsection{Texture Generation}
\label{subsec:texture-eval}


Our texture pipeline adds two design choices to standard differentiable rendering~\cite{Laine2020diffrast}: face alignment and super-resolution upscaling. \new{Because personalized appearance is a supporting component rather than a core contribution, we evaluate these as an internal ablation of our own design, not a benchmark against dedicated avatar methods.} We ablate both on the People Snapshot dataset~\cite{alldieck2018video} (four sequences, male/female-3/4-casual), holding out one-third of frames for testing with SMPL-X parameters from PyMAF-X rather than the dataset's, to simulate real-world conditions. Learned textures are rendered onto the test poses and compared to ground-truth RGB using three standard metrics: PSNR, SSIM, and LPIPS~\cite{zhang2018perceptual}.

\begin{table}[t]
\caption{Ablation study on face alignment and upscaling.}
\centering
\begingroup
\renewcommand{\arraystretch}{1.05}
\begin{tabularx}{\columnwidth}{l|*{3}{>{\centering\arraybackslash}X}}
\toprule
Condition & PSNR$\uparrow$ & SSIM$\uparrow$ & LPIPS$\downarrow$ \\
\midrule
Baseline & 25.289 & 0.9456 & 0.0644 \\
+ Face align & 25.289 & 0.9456 & 0.0640 \\
+ Upscaling & 25.291 & 0.9473 & 0.0639 \\
\rowcolor{olive!15}
+ Both & \textbf{25.294} & \textbf{0.9478} & \textbf{0.0630} \\
\bottomrule
\end{tabularx}
\endgroup
\label{tab:ablation_texture}
\end{table}

\begin{figure}[t]
  \centering
  \includegraphics[width=1\linewidth]{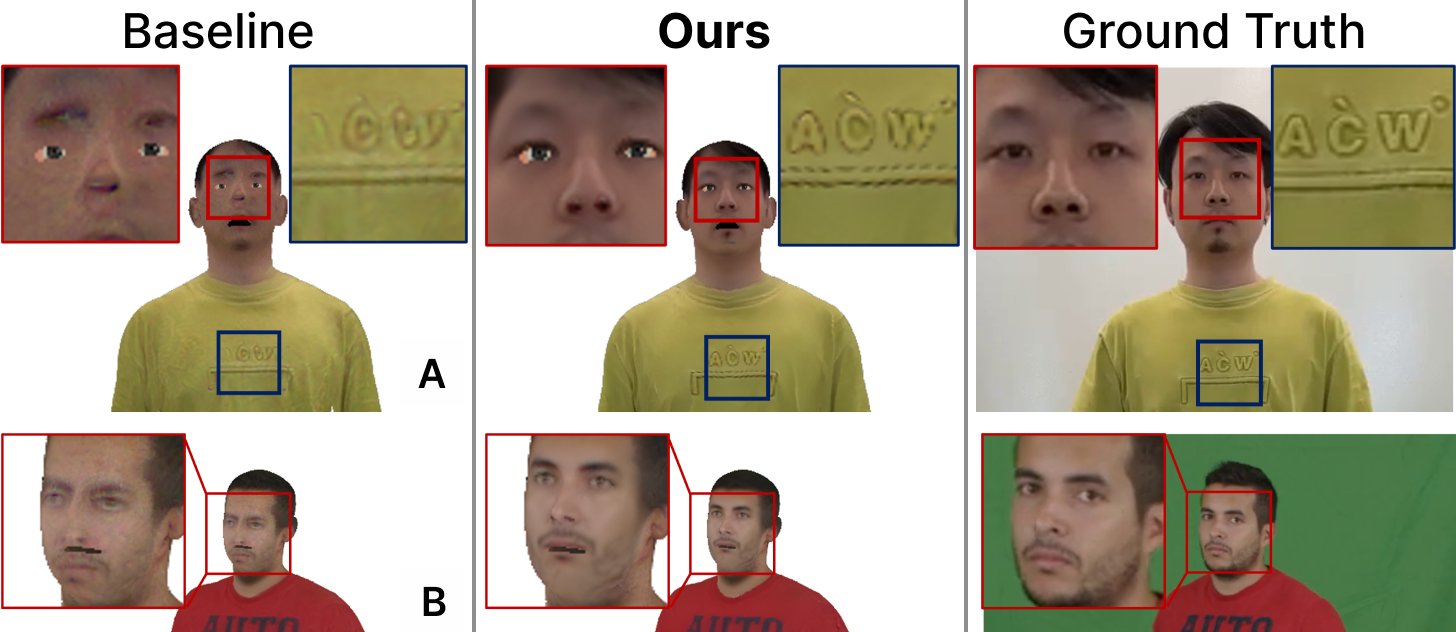}
  \caption{Texture ablation on (A) our capture data and (B) People Snapshot dataset. Face alignment and upscaling improve facial detail.}
  \label{fig:ablation_texture}
\end{figure}

\Cref{tab:ablation_texture} shows both steps help incrementally. \new{The gains are small only because they average over the full frame, where the face is a small share of pixels; on the face itself the improvement is clearly visible} (\Cref{fig:ablation_texture}): face alignment corrects eye and mouth registration errors from monocular pose, and upscaling recovers detail lost at $512\times512$. \new{With the pose model and texture pipeline validated, we next examine how these components shape the user's experience.}

\section{User Studies}
\label{sec:user-studies}

Having established that FlowAvatar integrates body, hand, face, and eye tracking into a single runtime with competitive, temporally stable pose, we now ask which of these components actually shape the avatar experience, a comparison that prior siloed systems could not support. We investigate two research questions:

\begin{itemize}
\setlength{\itemsep}{2pt}
\item \new{\textbf{RQ1}: does tracking completeness or personalized appearance more strongly shape \emph{third-person} perceived motion quality?}
\item \new{\textbf{RQ2}: how do tracking completeness and personalized appearance each shape \emph{first-person} embodiment (ownership, agency, and self-location)?}
\end{itemize}

We address RQ1 with a third-person motion quality study~(\Cref{sec:thirdperson}) and RQ2 with a first-person embodiment study~(\Cref{sec:firstperson}), each isolating tracking completeness from personalized look-alike texture. \new{Both run in the streaming configuration to log per-frame data under constant conditions.}

\begin{figure*}
  \centering
  \includegraphics[width=\textwidth]{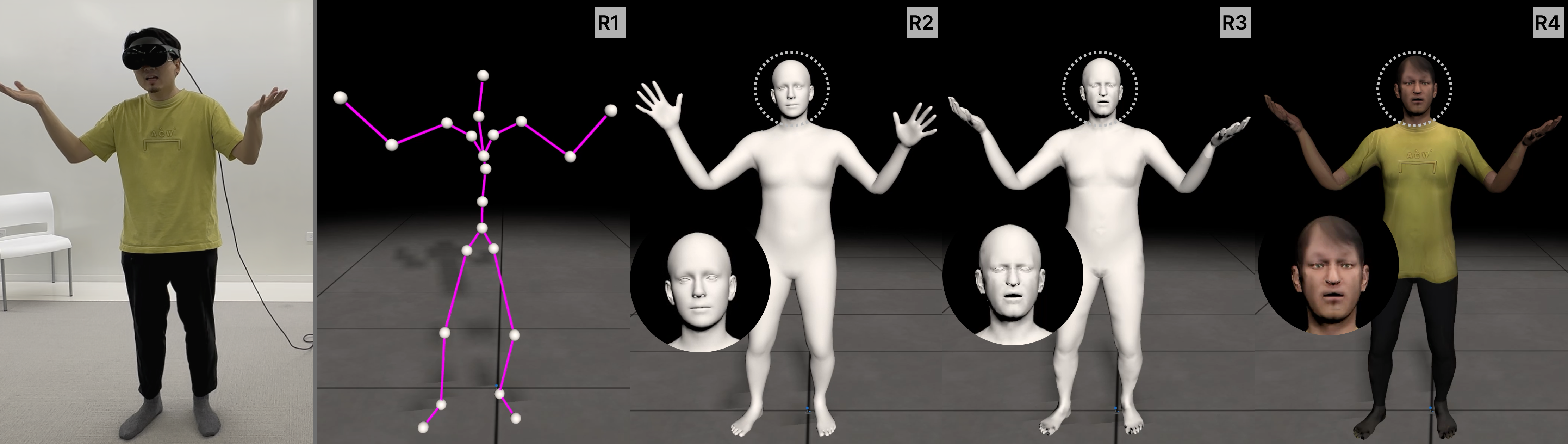}
  \caption{Example sequence (Anger) from the third-person motion quality study: ground truth, then representations (R1) Stick Figure, (R2) Minimal Body (no hands/face), (R3) Base Body (no texture), and (R4) Photo-real (personalized look-alike texture).}
  \label{fig:agency-study}
\end{figure*}

\subsection{Third-Person Motion Quality Study (N=12)}
\label{sec:thirdperson}

We examined how four avatar representations shape third-person perception of motion quality.

\subsubsection{Participants and Procedure}

We recruited 12 participants (7M/5F, aged 21--32, $M=25.17$, $SD=3.72$) from our university community, all with prior XR experience. \new{To hold motion identical across representations, an experimenter performed each sequence once on a Quest Pro in the streaming configuration. We logged the raw head and hand tracking input together with the model's predicted full-body pose, then re-rendered this single recording onto all four representations.} Participants viewed the resulting clips (\Cref{fig:agency-study}) in randomized order and rated each on six 7-point Likert dimensions ($-3$ to $+3$, adapted from~\cite{tao2023embodying}). Sequences and dimensions are listed in \Cref{tab:study-design}.

\begin{table}[t]
\caption{Motion sequences and evaluation dimensions (the four avatar representations R1--R4 are shown in \Cref{fig:agency-study}).}
\label{tab:study-design}
\centering
\small
\renewcommand{\arraystretch}{1.05}
\begin{tabularx}{\linewidth}{lX}
\toprule
\multicolumn{2}{l}{\textbf{Motion Sequences}} \\
\midrule
\textbf{S1: Greeting}   & Waving, smiling, clapping --- upper body gestures in social interaction \\
\textbf{S2: Anger}      & Frustrated gestures, angry expressions, assertive locomotion --- emotional expressivity \\
\textbf{S3: Joy}        & Jumping, smiling, rhythmic movement --- whole-body motion with subtle expressions \\
\textbf{S4: Exercise}   & Squatting, jumping, boxing --- dynamic full-body tracking under high intensity \\
\textbf{S5: Workplace}  & Sitting/standing transitions, desktop tasks --- fine hand movements and body transitions \\
\midrule
\multicolumn{2}{l}{\textbf{Evaluation Dimensions}} \\
\midrule
\textbf{Q1} & \textbf{Style \& Character:} How accurately does this representation capture the style, character, and expression of the original motion? \\
\textbf{Q2} & \textbf{Motion Fidelity:} How accurately does this representation capture the overall movement qualities? \\
\textbf{Q3} & \textbf{Communication:} How effectively could you understand what the person was trying to communicate? \\
\textbf{Q4} & \textbf{Interaction Readiness:} How confidently could you interact with this avatar based on its movements and expressions? \\
\textbf{Q5} & \textbf{Upper Body:} How effectively does this representation capture upper body movements (head, torso, arms, hands)? \\
\textbf{Q6} & \textbf{Lower Body:} How effectively does this representation capture lower body movements (hips, legs, feet)? \\
\bottomrule
\end{tabularx}
\end{table}

\begin{table}[h]
\caption{Motion quality: paired Wilcoxon results for the two key contrasts across Q1--Q6. $r$ is the rank-biserial effect size, CI the 95\% bootstrap CI of the paired mean difference. Exact-$p$ markers: $^{\dagger}p{<}.10$, $^{*}p{<}.05$, $^{**}p{<}.01$, $^{***}p{<}.001$; bold $p_{adj}$ survives Holm correction.}
\label{tab:mq-stats}
\centering\scriptsize\setlength{\tabcolsep}{2pt}
\begin{tabular}{lcccc}
\toprule
 & \multicolumn{2}{c}{\textbf{+Tracking (R2$\to$R3)}} & \multicolumn{2}{c}{\textbf{+Texture (R3$\to$R4)}} \\
\cmidrule(lr){2-3}\cmidrule(lr){4-5}
 & $p$ ($p_{adj}$) & $r$ [CI] & $p$ ($p_{adj}$) & $r$ [CI] \\
\midrule
Q1 & $\textless.001^{***}$ (\textbf{.004}) & +0.97 [+0.5, +1.1] & $.098^{\dagger}$ (.488) & +0.58 [$-$0.0, +0.8] \\
Q2 & $\textless.001^{***}$ (\textbf{.001}) & +1.00 [+0.6, +1.5] & $.512$ (.948) & +0.25 [$-$0.2, +0.4] \\
Q3 & $\textless.001^{***}$ (\textbf{.001}) & +1.00 [+0.8, +1.7] & $.316$ (.948) & +0.33 [$-$0.1, +0.5] \\
Q4 & $.014^{*}$ (\textbf{.014}) & +0.85 [+0.3, +1.1] & $.004^{**}$ (\textbf{.023}) & +0.96 [+0.3, +1.0] \\
Q5 & $.002^{**}$ (\textbf{.004}) & +1.00 [+0.7, +1.8] & $.152$ (.609) & +0.56 [$-$0.0, +0.3] \\
Q6 & $\textless.001^{***}$ (\textbf{.004}) & +1.00 [+0.6, +1.3] & $.439$ (.948) & +0.29 [$-$0.2, +0.5] \\
\bottomrule
\end{tabular}
\end{table}

\begin{figure}[h]
  \centering
  \includegraphics[width=1\linewidth]{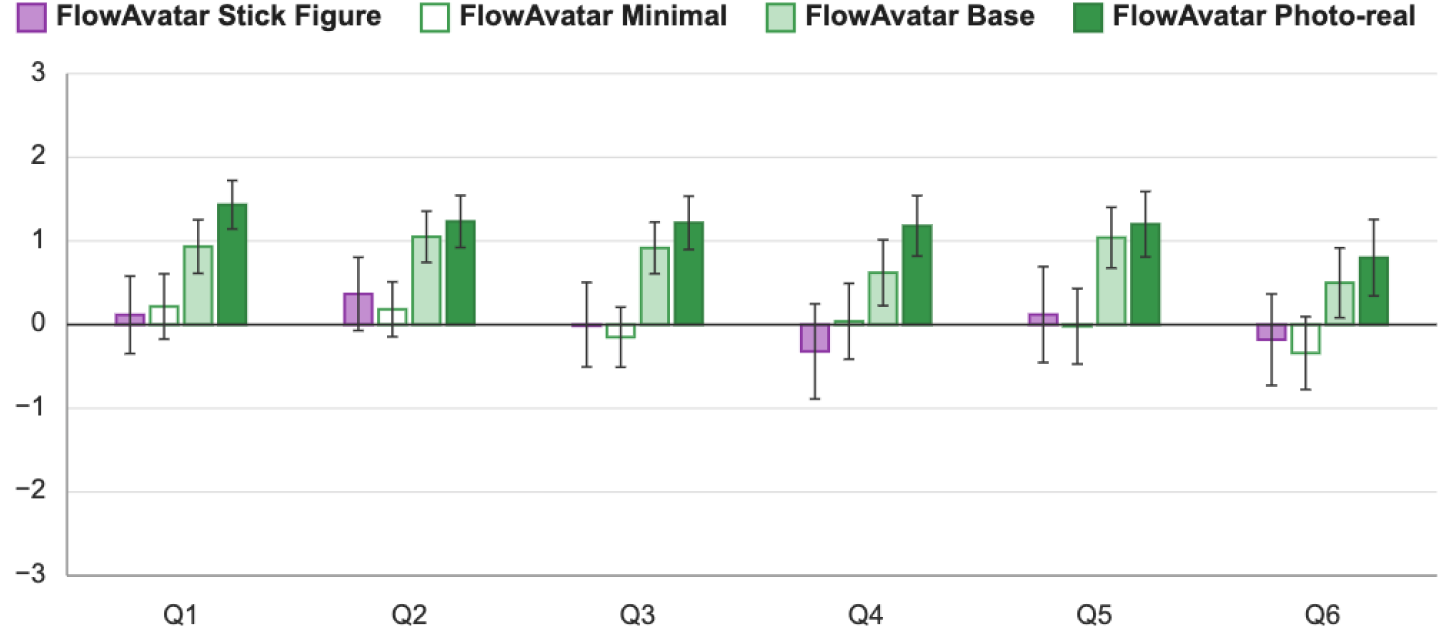}
  \caption{Third-person motion quality study: Mean ratings with standard error for four avatar representations across five motion sequences and six evaluation dimensions.}
  \label{fig:Third-Person}
\end{figure}

\subsubsection{Results}

\Cref{fig:Third-Person} shows mean ratings for the four representations across the six evaluation dimensions. Photo-real (R4) received the highest overall ratings ($M=1.19$, $SD=1.09$), followed by Base Body (R3; $M=0.92$, $SD=1.11$), while Minimal Body (R2; $M=-0.07$, $SD=1.27$) and Stick Figure (R1; $M=0.00$, $SD=1.60$) scored substantially lower. Friedman tests confirmed significant main effects on all six dimensions (all $p < 0.003$, Kendall's $W$ 0.42--0.54), despite identical underlying motion. \new{Adding body geometry alone (R1$\to$R2) did not improve ratings: a body mesh without articulated face and hand tracking conveys little beyond a skeleton, so we focus the contrasts on tracking completeness (R2$\to$R3) and texture (R3$\to$R4).}

We compared conditions with Wilcoxon signed-rank tests, Holm-corrected across the six dimensions (\Cref{tab:mq-stats}). Adding face and hand tracking (R2$\to$R3) improved all six dimensions (all $p_{adj} \leq 0.014$) with large, consistent effect sizes (rank-biserial $r \geq 0.85$, all CIs excluding zero), whereas adding personalized texture (R3$\to$R4) reached significance only for Interaction Readiness (Q4, $p_{adj} = 0.023$).

\subsection{First-Person Embodiment Study (N=10)}
\label{sec:firstperson}

We assessed FlowAvatar's effectiveness in creating first-person embodiment, specifically the sense of ownership and agency over one's virtual avatar.

\subsubsection{Participants and Procedure}

\begin{figure}[t]
  \centering
  \includegraphics[width=1\linewidth]{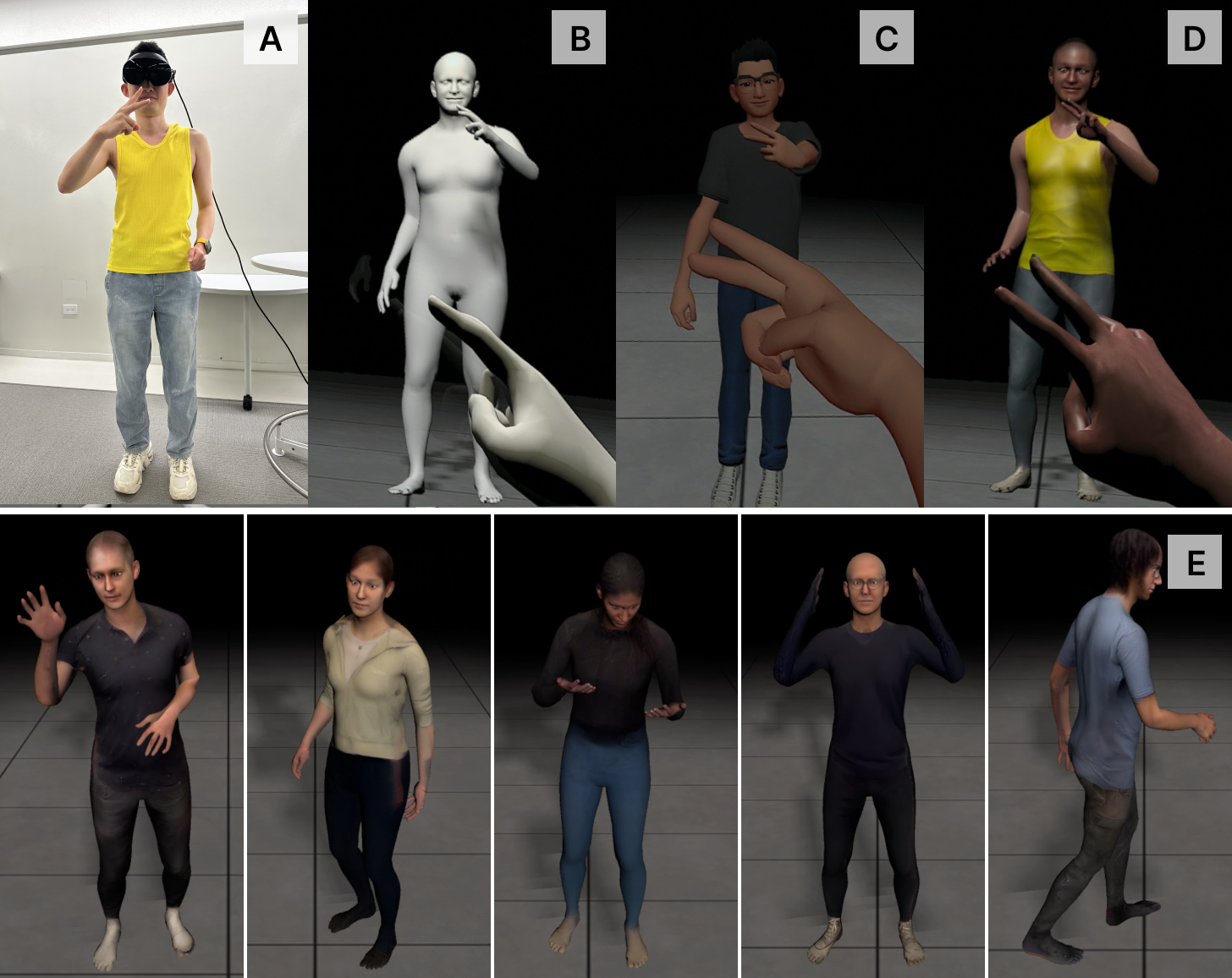}
  \caption{Embodiment study: (A) participant's pose; avatar representations: (B) FlowAvatar Base Body; (C) Meta Avatar; (D) FlowAvatar Photo-real (personalized look-alike texture). (E) Five participants' avatars during study.}
  \label{fig:embodiment-users}
\end{figure}

We recruited 10 new participants (6M/4F, mean age 30.1, $SD=9.46$), all with prior XR experience. Sessions averaged 53 minutes. Each first created a personalized avatar with our web tool and a cartoon avatar in Meta Horizon~\cite{metaAvatars}. All conditions ran in the same Unity scene, a virtual mirror (\Cref{fig:embodiment-users}) in which participants performed the same free-form movements (walking, jumping, gestures, and facial expressions), so only the avatar differed: two were driven by FlowAvatar, Base Body (full tracking, no texture) and Photo-real (full system with personalized texture), while the third loaded the participant's Meta Horizon avatar (upper body only) after signing in to Meta. Conditions were shown in varying order across participants, with participants blind to the driving system. The Photo-real vs.\ Base Body comparison isolates texture with tracking held constant, whereas Photo-real vs.\ Meta Avatar additionally varies tracking and visual style, serving as an external consumer baseline rather than a controlled contrast. Participants rated each condition on the CoolMoves questionnaire~\cite{ahuja2021coolmoves,peck2021avatar}: ownership ($\pm$15), agency ($\pm$12), location ($\pm$9), appearance ($\pm$6), response ($\pm$6), and total embodiment ($\pm$3). \new{After each session, a $\sim$10-minute semi-structured interview was coded by two researchers into recurring themes (self-recognition, motion coherence, and tracking predictability), illustrated with representative quotes below.}

\subsubsection{Results}

\begin{figure}[h]
  \centering
  \includegraphics[width=1\linewidth]{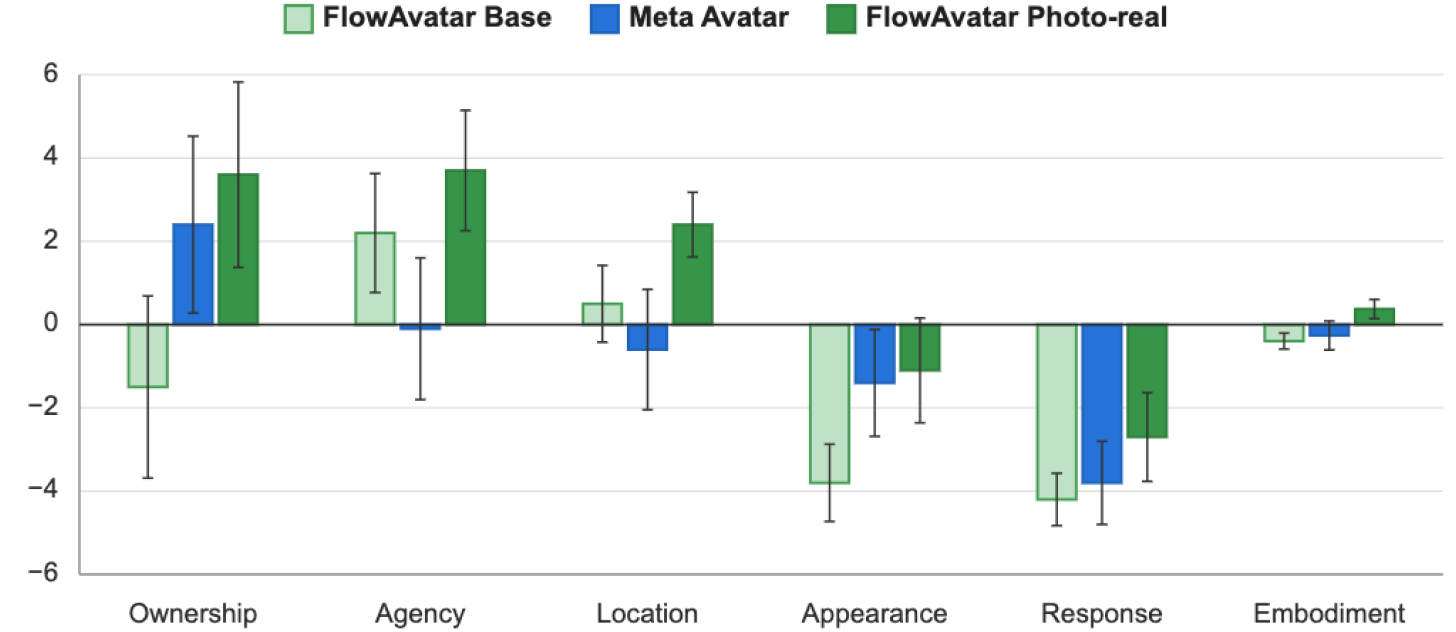}
  \caption{First-person embodiment study: CoolMoves~\cite{ahuja2021coolmoves,peck2021avatar} subscale scores (centered at 0) for three avatar conditions. Photo-real was the only condition with positive Total Embodiment.}
  \label{fig:First-Person}
\end{figure}

\Cref{fig:First-Person} shows CoolMoves~\cite{ahuja2021coolmoves,peck2021avatar} subscale scores across all three conditions. Photo-real scored numerically highest on all six dimensions and was the only condition achieving positive total embodiment. A Friedman test showed a significant main effect for Appearance ($\chi^2(2)=6.74$, $p=0.034$, $W=0.337$) and a marginal trend for Agency ($\chi^2(2)=5.03$, $p=0.081$).

\begin{table}[h]
\caption{Embodiment (N=10, exploratory): paired Wilcoxon results. Photo vs.\ Base isolates texture (tracking held constant). Photo vs.\ Meta additionally varies tracking and style. $r$ is rank-biserial, CI the 95\% bootstrap CI of the paired mean difference. Exact-$p$ markers: $^{\dagger}p{<}.10$, $^{*}p{<}.05$; no comparison survives Holm correction ($p_{adj}$).}
\label{tab:emb-stats}
\centering\scriptsize\setlength{\tabcolsep}{0.6pt}
\begin{tabular}{@{}lcccc@{}}
\toprule
 & \multicolumn{2}{c}{\textbf{Photo vs.\ Base}} & \multicolumn{2}{c}{\textbf{Photo vs.\ Meta}} \\
\cmidrule(lr){2-3}\cmidrule(lr){4-5}
Subscale & $p$\,($p_{adj}$) & $r$ [CI] & $p$\,($p_{adj}$) & $r$ [CI] \\
\midrule
Ownership & $.062^{\dagger}$\,(.250) & +0.67 [+0.7,\,+9.2] & $.668$\,(1.000) & +0.18 [$-$4.6,\,+6.6] \\
Agency & $.125$\,(.250) & +0.64 [+0.0,\,+3.1] & $.043^{*}$\,(.258) & +0.73 [+0.8,\,+7.3] \\
Location & $.102$\,(.250) & +0.67 [+0.1,\,+3.9] & $.109$\,(.547) & +0.64 [+0.2,\,+6.4] \\
Appearance & $.031^{*}$\,(.164) & +1.00 [+0.9,\,+5.0] & $1.000$\,(1.000) & $-$0.04 [$-$1.3,\,+2.4] \\
Response & $.078^{\dagger}$\,(.250) & +0.82 [+0.3,\,+3.0] & $.500$\,(1.000) & +0.67 [$-$0.2,\,+2.8] \\
Total & $.027^{*}$\,(.164) & +0.78 [+0.3,\,+1.2] & $.375$\,(1.000) & +0.35 [$-$0.0,\,+1.5] \\
\bottomrule
\end{tabular}
\end{table}

\paragraph{Ownership.}
Photo-real achieved the highest ownership scores ($M=+3.60$, $SD=7.04$), with participants noting personal recognition: ``Since FlowAvatar has the same outfit as mine, I think FlowAvatar is my avatar'' [P-004]. Meta Avatar showed moderate ownership ($M=+2.40$, $SD=6.72$), though participants found it unconvincing: ``I don't think the cartoon looks like me'' [P-002]. Base Body scored lowest ($M=-1.50$, $SD=6.92$), lacking personalization despite identical motion to Photo-real. The Photo-real vs.\ Base Body difference showed a marginal trend ($p=0.062$).

\paragraph{Agency.}
Photo-real ($M=+3.70$, $SD=4.57$) substantially exceeded Meta Avatar ($M=-0.10$, $SD=5.38$), reaching significance ($p=0.043$). Participants specifically mentioned motion capabilities: ``What surprised me is that the FlowAvatar's jump and squat are nice'' [P-006]. A key differentiator was tracking behavior: ``When I raise my hands up...Meta ones will fall down, which makes me feel scary because I did not do that'' [P-007], highlighting how FlowAvatar's approach to handling tracking loss created a more predictable experience. Base Body scored moderately ($M=+2.20$, $SD=4.52$). Meta Avatar's lack of lower-body animation was a significant limitation for agency.

\paragraph{Location.}
Photo-real scored highest for body location ($M=+2.40$, $SD=2.46$), with integrated tracking creating spatial presence: ``All my tracking, hands, face are pretty nice in FlowAvatar, with these all together, I can see in mirror it was me'' [P-008]. Base Body received moderate positive scores ($M=+0.50$, $SD=2.92$), as one participant noted ``I know the white body is not me, it only has my motion data'' [P-010], while Meta Avatar received lower location scores ($M=-0.60$, $SD=4.58$), with participants reporting disconnection between their physical movements and the avatar's representation.

\paragraph{Appearance and Response.}
Photo-real ($M=-1.10$, $SD=3.98$) scored higher than Base Body ($M=-3.80$, $SD=2.94$) on Appearance (uncorrected $p=0.031$, Holm-corrected $p_{adj}=0.164$), the only dimension with a significant Friedman main effect. Meta Avatar scored between the two ($M=-1.40$, $SD=4.06$). All three conditions received negative appearance scores. For Response, Photo-real scored highest ($M=-2.70$, $SD=3.37$) followed by Meta Avatar ($M=-3.80$, $SD=3.16$) and Base Body ($M=-4.20$, $SD=1.99$), with a marginal Photo-real vs.\ Base Body trend ($p=0.078$). Participants generally did not feel their real bodies were affected by virtual avatar interactions.

\paragraph{Total Embodiment.}
Photo-real ($M=+0.37$, $SD=0.73$) was the only condition achieving positive total embodiment, scoring higher than Base Body ($M=-0.40$, $SD=0.61$, uncorrected $p=0.027$, $p_{adj}=0.164$). Meta Avatar ($M=-0.26$, $SD=1.10$) also scored below neutral.

\new{We also compared conditions pairwise with Wilcoxon signed-rank tests, Holm-corrected across the six subscales and treated as exploratory (\Cref{tab:emb-stats}). The Photo-real vs.\ Base Body comparison (isolating texture) reached significance on Appearance ($p=.031$) and Total Embodiment ($p=.027$), with marginal trends on Ownership ($p=.062$) and Response ($p=.078$), whereas the Photo-real vs.\ Meta Avatar comparison reached significance only on Agency ($p=.043$). These effects carried large effect sizes (rank-biserial $r \geq 0.64$), but none survived Holm correction at $N=10$.}

\section{Discussion}
\label{sec:discussion}

\new{Returning to our two research questions, both studies point to a consistent division of labor between tracking and appearance.} In the motion quality study (\textbf{RQ1}), adding face and hand tracking improved all six evaluation dimensions (Holm-corrected, all $p_{adj} \leq 0.014$), whereas adding texture reached significance on only one, so tracking completeness dominates third-person motion quality. The effect was strongest for emotional content: in the Anger sequence, face/hand tracking raised Style and Character scores from $M=0.00$--$0.08$ to $M=1.38$--$1.58$, indicating that articulated hands and faces carry emotional intent that body pose alone conveys poorly. In the embodiment study (\textbf{RQ2}), FlowAvatar's tracking was associated with greater agency than Meta Avatar ($p=.043$, uncorrected) and texture with ownership and self-recognition. \new{We read the agency effect as suggestive, given the confounded Meta comparison.} Participants consistently valued motion coherence, as ``For the avatar movement, it has more natural movement'' [P-002], and predictable tracking behavior, with P-007 noting that Meta Avatar's hands would unexpectedly drop when raised overhead.

This user-perceived quality follows from the model's design. The comprehensive temporal losses (\Cref{subsec:losses}) are the single most impactful loss category in our ablation, reducing jitter by 37\% (\Cref{tab:ablation_loss}), and this temporal consistency plausibly underlies the reliable agency participants reported: predictable full-body behavior mattered more for felt agency than visual fidelity. More broadly, for sparse egocentric tracking, pairing comprehensive spatial and temporal losses with a lightweight recurrent architecture yields the best temporal consistency in our benchmark (\Cref{tab:ablation_backbone}).

Personalized texture serves a complementary but distinct role. It scored higher on Appearance ($p=.031$) and Total Embodiment ($p=.027$) than Base Body, \new{though neither effect survives Holm correction at this small sample}. Participants valued self-recognition: ``I prefer the photorealistic one, compared to the cartoon one, it looks more like me'' [P-009]. However, uncanny valley effects emerged: ``It is like someone with your clothes but looks a bit different than you, but still, it looks pretty close to me already'' [P-003]. Several participants preferred cartoon avatars in social contexts: ``I still think virtual world should be something virtual, you can't bring the `real' me there'' [P-010]. We also observed that current limitations in hair modeling and facial geometry may affect avatar quality unevenly across users, as ``it can't have better solution to show my hair, which makes me ugly'' [P-004]. These reactions suggest that the value of personalized look-alike texture depends on application context and avatar quality, and that equitable representation remains an open challenge. Practically, when compute-constrained, \new{these preliminary findings suggest} prioritizing tracking over rendering fidelity: \new{in our studies,} complete tracking of body, face, and hands had the broadest impact on motion quality and agency, while texture added value mainly through self-recognition and ownership. Notably, these interactions between tracking and appearance only become visible through integrated evaluation, reinforcing the value of holistic system assessment over optimizing avatar subsystems independently.

\section{Limitations and Future Work}

Our pose estimation generalizes well across diverse movements, though it occasionally faces challenges with poses not well-represented in training data. The current implementation treats environmental interactions as soft constraints through loss functions rather than hard physical constraints, achieving plausible results at runtime. Future work could extend beyond flat ground planes to include full-body environment contact~\cite{winkler2022questsim}, object interactions~\cite{brahmbhatt2020contactpose}, and additional sensor fusion (e.g., IMUs) for improved lower-body estimation.

The SMPL-X mesh constrains visual fidelity: Unity's implementation reduces the blendshape space from 100 to 10 expressions, and the model cannot represent hair, clothing dynamics, or changing lighting, with no separation between skin and fabric. \new{External-sensor capture would reach higher photorealism, a fidelity we deliberately trade for a smartphone-only, consumer-accessible enrollment pipeline.} Hybrid approaches with codec avatars~\cite{martinez2024codec} or dynamic rendering~\cite{saito2024relightable}, and emerging Gaussian Splatting~\cite{hu2024gaussianavatar}, could address these. Recent mobile-ready Gaussian avatar distillation~\cite{iandola2025squeezeme} suggests such upgrades are becoming feasible on consumer hardware.

Our on-device configuration was not evaluated perceptually. Future work should assess whether its quality difference from the streaming configuration, and the tracker-bound 97~ms motion delay (\Cref{sec:deployment}), are noticeable in practice.

\new{The user studies are preliminary. With modest samples (N=12, N=10) and no a-priori power analysis, we report effect sizes and confidence intervals throughout, and several embodiment effects (such as Photo-real over Base Body on Appearance and Total Embodiment) hold only before multiple-comparison correction. The Meta Avatar condition is an external baseline, not a controlled contrast: its closed interface varies tracking, rendering, and lower-body animation together. In the motion quality study, the Q1 item bundled style, character, and expression into one rating, limiting per-component attribution, though the tracking effect held across all six items. Larger, appropriately powered studies are needed to confirm these effects.}

\new{Finally, FlowAvatar integrates multimodal motion signals (head, hand, face, gaze, full-body pose) that can reveal sensitive personal attributes, with risks compounding across modalities~\cite{nair2024inferring}. Our on-device runtime mitigates this by keeping raw motion on the headset rather than streaming it; privacy-preserving representation and storage remain important future work.}

\section{Conclusion}

FlowAvatar demonstrates that a unified avatar system connecting personalized look-alike appearance generation to real-time full-body pose, hand articulation, facial expression, and eye gaze tracking is achievable on a standalone consumer headset without additional instrumentation. A pose estimation framework combining body-part encoding, unified kinematic decoding, and comprehensive spatial and temporal losses achieves competitive spatial accuracy with state-of-the-art temporal consistency on the AMASS benchmark for this sparse input regime. Two user studies provide converging evidence that tracking completeness and personalized appearance serve complementary roles: tracking completeness drives perceived motion quality and is associated with agency, personalized texture with ownership and self-recognition, and only their combination achieves positive total embodiment. We release the complete pipeline as open source at \url{https://github.com/SPICExLAB/FlowAvatar}, lowering the barrier to embodied avatars on consumer hardware and inviting the community to build on these findings.

\bibliographystyle{abbrv-doi-narrow}
\bibliography{bibtex}
\end{document}